\documentclass[runningheads,a4paper]{llncs}

\usepackage{amssymb}
\usepackage{graphicx}
\usepackage[utf8]{inputenc}
\usepackage{url}
\usepackage{tabularx}
\usepackage{epstopdf}
\usepackage{graphicx}
\usepackage{caption}
\usepackage{amsmath}
\usepackage{subcaption}
\usepackage{subfig} 
\usepackage{float}
\usepackage[parfill]{parskip}

\urldef{\mailsa}\path|{pplonski, zaremba}@ire.pw.edu.pl|    
\newcommand{\keywords}[1]{\par\addvspace\baselineskip
\noindent\keywordname\enspace\ignorespaces#1}

\begin{document}

\mainmatter 
\title{Electron Neutrino Classification in Liquid Argon Time Projection Chamber Detector} 
\titlerunning{$\nu_e$ Classification in LAr-TPC Detector}
\author{Piotr Płoński$^1$, Dorota Stefan$^2$, Robert Sulej$^3$, Krzysztof Zaremba$^1$}
\authorrunning{P. Płoński et al.}
\institute{$^1$Institute of Radioelectronics, Warsaw University of Technology,\\ Nowowiejska 15/19,00-665 Warsaw, Poland,\\
\mailsa
\\
$^2$Instituto Nazionale di Fisica Nucleare, Sezione di Milano e Politecnico,\\Via Celoria 16, I-20133 Milano, Italy
\\\path|dorota.stefan@ifj.edu.pl|
\\
$^3$National Center for Nuclear Research,\\A. Soltana 7, 05-400 Otwock/Swierk, Poland
\\
\path|Robert.Sulej@cern.ch| 
}
\maketitle

\begin{abstract}

Neutrinos are one of the least known elementary particles. The detection of neutrinos is an extremely difficult task since they are affected only by weak sub-atomic force or gravity. Therefore large detectors are constructed to reveal neutrino's properties. Among them the Liquid Argon Time Projection Chamber (LAr-TPC) detectors provide excellent imaging and particle identification ability for studying neutrinos. The computerized methods for automatic reconstruction and identification of particles are needed to fully exploit the potential of the LAr-TPC technique. Herein, the novel method for electron neutrino classification is presented. The method constructs a feature descriptor from images of observed event. It characterizes the signal distribution propagated from vertex of interest, where the particle interacts with the detector medium. The classifier is learned with a constructed feature descriptor to decide whether the images represent the electron neutrino or cascade produced by photons. The proposed approach assumes that the position of primary interaction vertex is known. The method's performance in dependency to the noise in a primary vertex position and deposited energy of particles is studied. 

\keywords{Electron Neutrino, Classification, Image Descriptor, Liquid Argon, Time Projection Chambers
}

\end{abstract}

\section{Introduction}

Neutrinos are one of the fundamental particles as well as one of the least understood. They exist in three flavors: electron, muon and tau. There is a hypothesis about the existence of the fourth type of flavor, namely sterile \cite{sterile}. The detection of neutrinos is an extremely difficult task since they are affected only by weak sub-atomic force or gravity. Therefore, large detectors are constructed to reveal neutrino's properties. Among them the Liquid Argon Time Projection Chamber (LAr-TPC) detector, proposed by C.Rubbia in 1977 \cite{Rubbia}, provides excellent imaging ability of charged particles, making it ideal for studying neutrino oscillation parameters, sterile neutrinos existence \cite{sterile}, Charge Parity violation, violation of baryonic number conservation and dark matter searches. The LAr-TPC technique is used in several projects around the world \cite{microboone}, \cite{maly_rubbia},  \cite{Argoneut}, \cite{lbne}, \cite{design-t600}. Among them, the ICARUS T600 \cite{design-t600} was the largest working detector located at Gran Sasso in underground Italian National Laboratory operating on CNGS beam (CERN\footnote{CERN - European Organization for Nuclear Research} Neutrinos to Gran Sasso). In this study, the T600 parameters will be used since other existing or planned LAr-TPC detectors have the same or similar construction and settings.  

A neutrino which passes through the LAr-TPC detector, can interact with nuclei of argon. Density of argon in liquid form makes the rate of interactions practical for experimental study. Charged particles, created in interaction, produce both scintillation light and ionization electrons along its path in the LAr-TPC detector. The scintillation light, which is poor compared to ionisation charge, is detected by photomultipliers which trigger the read-out process. Free electrons from ionizing particles drift in a highly purified liquid argon in an uniform electric field toward the anode. Electrons diffussion approximate value 4.8 cm$^2$/s is much slower than electron drift velocity 1.59 mm/$\mu$s, therefore they can drift to macroscopic distances preserving high resolution of track details. The anode consists of three wire planes, so-called Induction1, Induction2, Collection. A signal is induced in a non-destructive way on the first two wire planes, which are practically transparent to the drifting electrons. The signal on the third wire plane (Collection) is formed by collecting the ionization charge. The wires in consecutive planes are oriented in three different degrees with respect to the horizontal with 3 mm spacing between wires in the plane. This allows to localize the signal source in the XZ plane, whereas Y coordinate is calculated from wire signal timing and electron drift velocity\footnote{Coordinate system labeling is given for reference.}. Signal on wires is amplified and digitized with 2.5 MHz sampling frequency which results in 0.64 mm spatial resolution along the drift coordinate. The digitized waveforms from consecutive wires placed next to each other form 2D projection images of an event, with resolution 0.64 mm x 3 mm.

One of the common aims of proposed and future neutrino experiments is to study the appearance of electron neutrinos in the muon neutrino beam. Fundamental requirement for a such study is the method for classification of the interacting neutrino flavour and, in the case of detectors placed on surface, the method for the cosmogenic background rejection. Selection of $\nu_e$ interaction among other $\nu$ interactions and background events should involve analysis of the primary interaction vertex (PIV) features, including detection of a single electron and presence of hadronic activity. These features may allow to eliminate the majority of events that can mimic signal, namely:
\begin{itemize}
\item $\nu$ interactions with $\pi_0$ produced in the vertex, which decays into gammas immediately, and one or more gammas converts to e+/e- in close vicinity of the vertex;
\item gammas produced by cosmogenic sources, converting in the detector volume within the data taking time window with production of e+/e- pair. 
\end{itemize}

In this paper, a novel method for automatic classification of $\nu_e$ from cosmogenic sources is presented. The considered range of energy deposited by an event in the detector is 0.2-1.0 GeV. We expect that appearance of interesting $\nu_e$ events within this range and therefore we are preparing the method for rejection of background events resulting in similar energy deposit. In the proposed method for each event a feature descriptor is constructed. It describes the signal distribution in Induction2 and Collection views. The method assumes that the localization of the PIV, where the particle starts interaction with detector, is known. The classifier is learned with a created feature descriptor. Herein, the different settings used in the feature vector construction are examined. The settings with the best performance are selected. The impact of noise in PIV localization on method's performance is analyzed. Additionally, the classifier performance is assessed on various energy ranges of classified events.

\section{Methods}

\subsection{Dataset creation}

The dataset was generated with the FLUKA software \cite{fluka} and T600 detector parameters. There were generated 7090 events with energy from 0.2 to 1.0 GeV equally distributed. Among them, there were 3283 events from electron neutrino (positive class label) and 3807 events from cosmogenic sources (negative class label). For each event the position of primary interaction vertex is assumed to be known. All the events have PIV located at least 5 cm from anode or cathode. All images were deconvoluted with impulse response of the wire signal readout chain. The segmentation procedure \cite{segmentationLArTPC} was applied to remove detector's noise from images. There are considered two views for each event, namely, Induction2 and Collection. From each view the image chunk with size 101 wires x 505 samples and center in the PIV was considered. The used chunk's size is sufficiently large for analysis conducted in this paper. The images where downsampled to 101 x 101 pixels size to provide similar resolution on both axis, where 1 wire corresponds to 1 pixel in the x-axis and 5 samples corresponds to 1 pixel in the y-axis. 

\subsection{Event's feature descriptor}

\begin{figure}[H]
\captionsetup[subfigure]{aboveskip=1pt,belowskip=1pt}
\centering
\begin{subfigure}[b]{0.28\textwidth}
	\caption{Raw, Ind2, bkg.}  
	\includegraphics[clip=true, width=\textwidth]{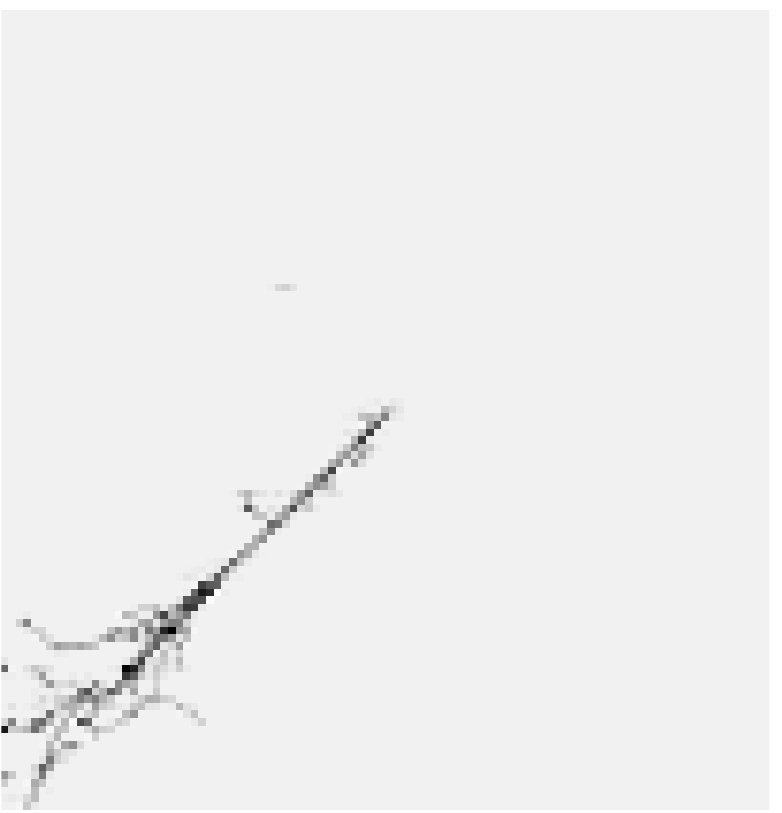}
\end{subfigure}
\begin{subfigure}[b]{0.298\textwidth}
	\caption{Polar, Ind2, bkg.}  
	\includegraphics[clip=true, width=\textwidth]{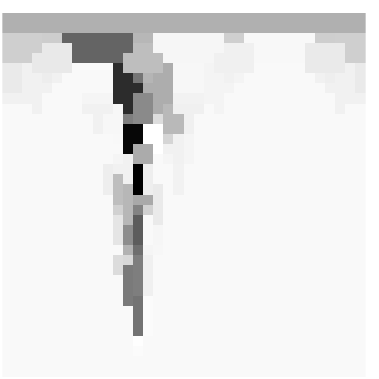}
\end{subfigure}
\begin{subfigure}[b]{0.39\textwidth}
	\caption{Distribution, Ind2, bkg.}  
	\includegraphics[clip=true, width=\textwidth]{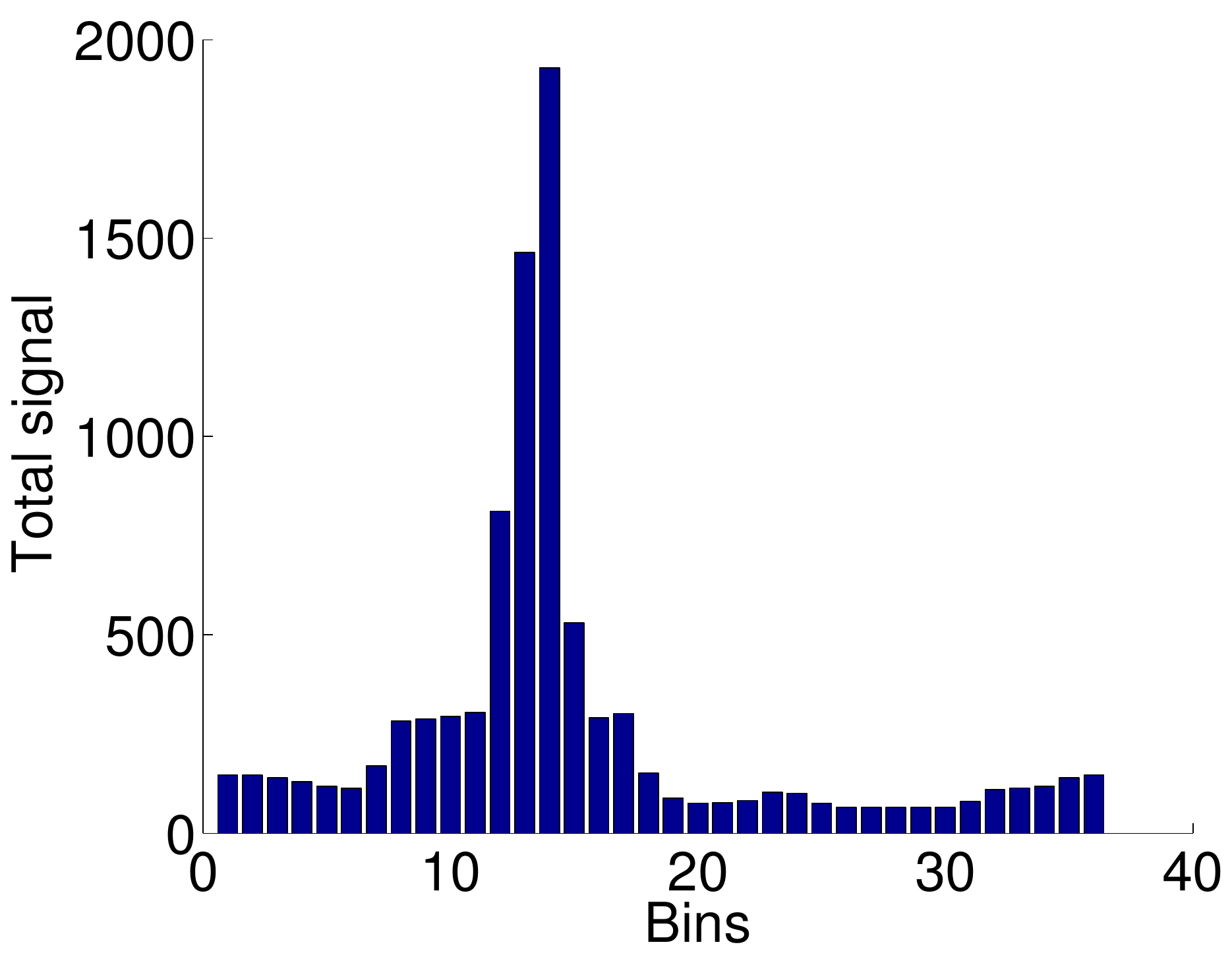}
\end{subfigure}
\begin{subfigure}[b]{0.28\textwidth}
	\caption{Raw, Coll, bkg.}  
	\includegraphics[clip=true, width=\textwidth]{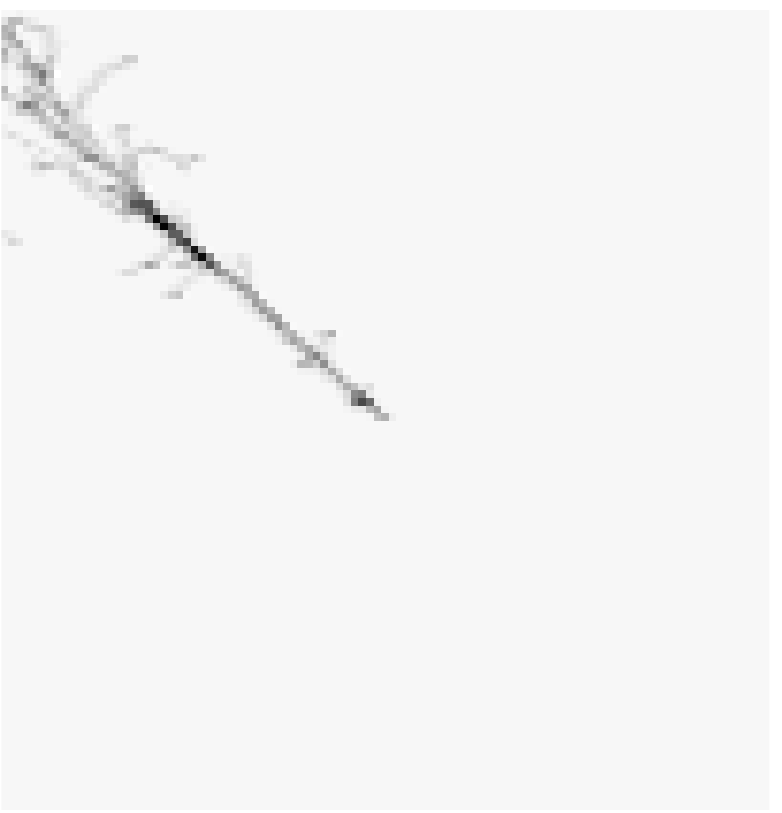}
\end{subfigure}
\begin{subfigure}[b]{0.298\textwidth}
	\caption{Polar, Coll, bkg.}  
	\includegraphics[clip=true, width=\textwidth]{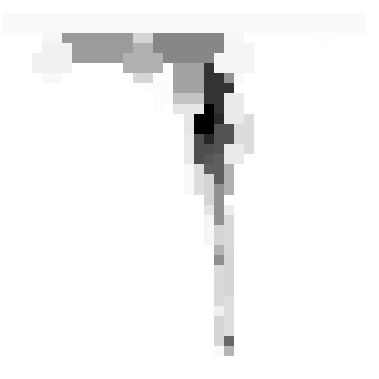}
\end{subfigure}
\begin{subfigure}[b]{0.39\textwidth}
	\caption{Distribution, Coll, bkg.}  
	\includegraphics[clip=true, width=\textwidth]{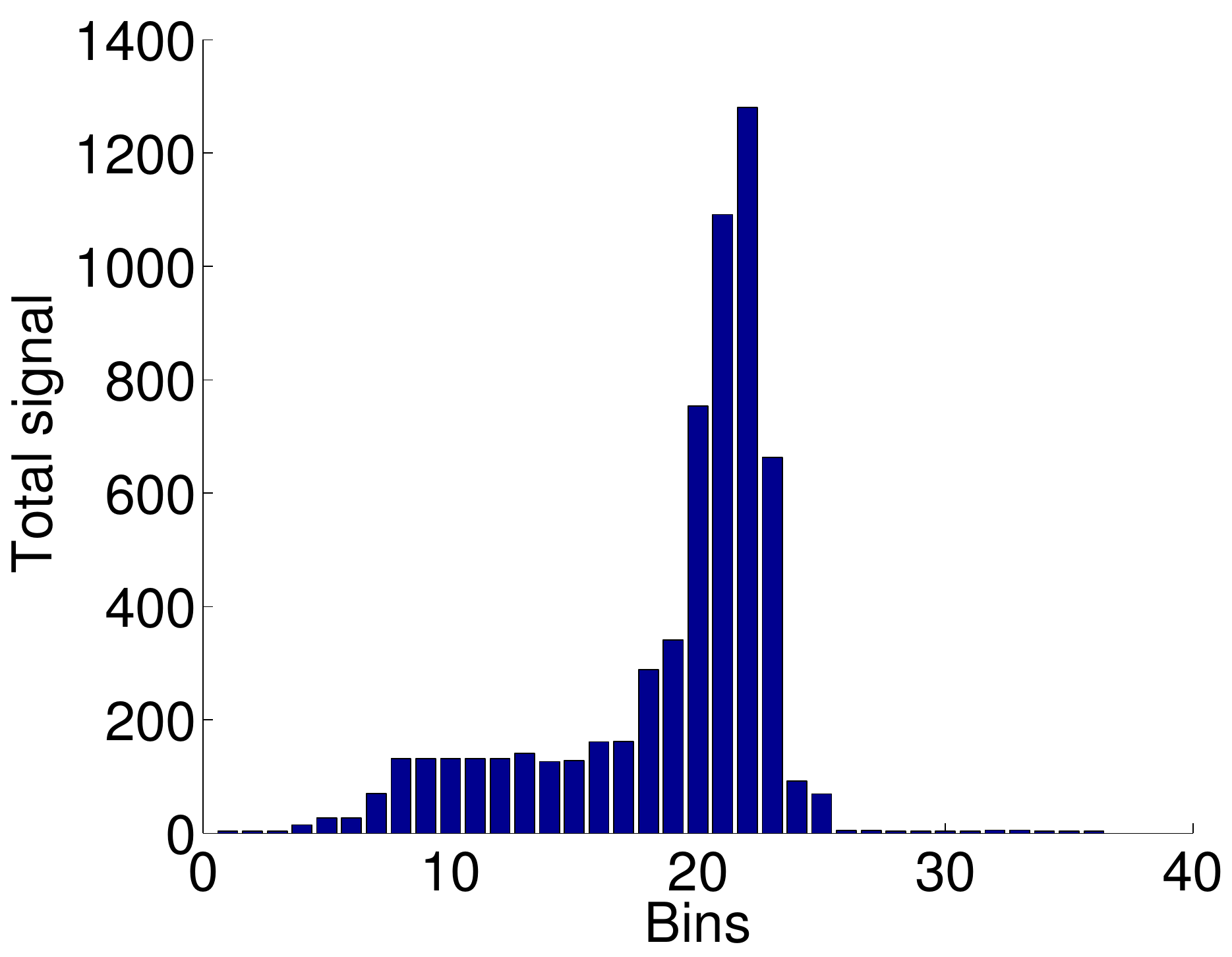}
\end{subfigure}
\begin{subfigure}[b]{0.28\textwidth}
	\caption{Raw, Ind2, sig.}  
	\includegraphics[clip=true, width=\textwidth]{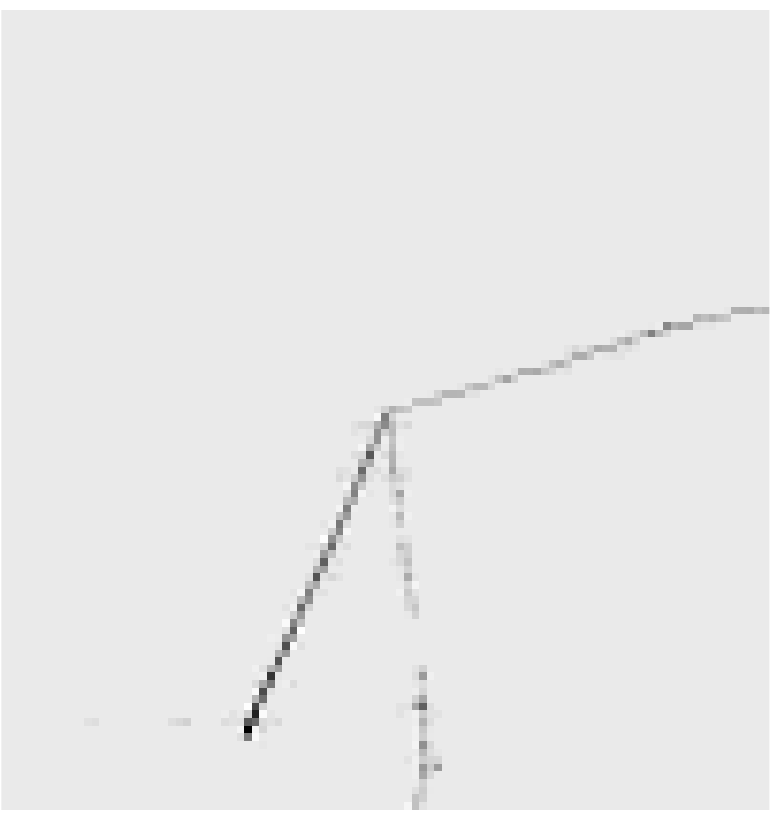}
\end{subfigure}
\begin{subfigure}[b]{0.298\textwidth}
	\caption{Polar, Ind2, sig.}  
	\includegraphics[clip=true, width=\textwidth]{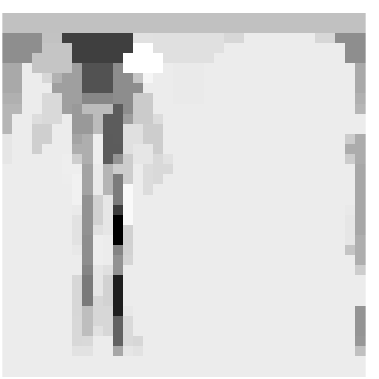}
\end{subfigure}
\begin{subfigure}[b]{0.39\textwidth}
	\caption{Distribution, Ind2, sig.}  
	\includegraphics[clip=true, width=\textwidth]{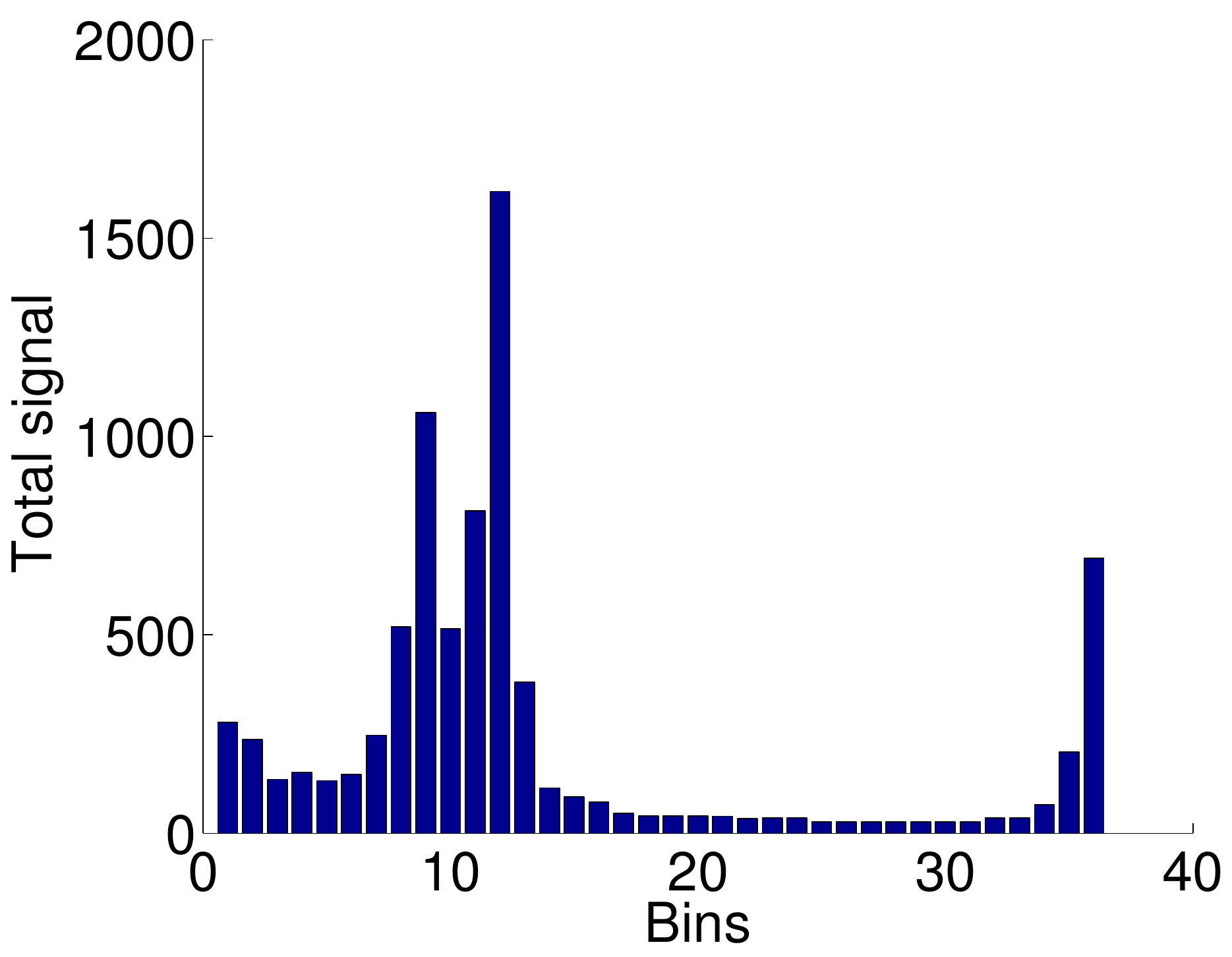}
\end{subfigure}
\begin{subfigure}[b]{0.28\textwidth}
	\caption{Raw, Coll, sig.}  
	\includegraphics[clip=true, width=\textwidth]{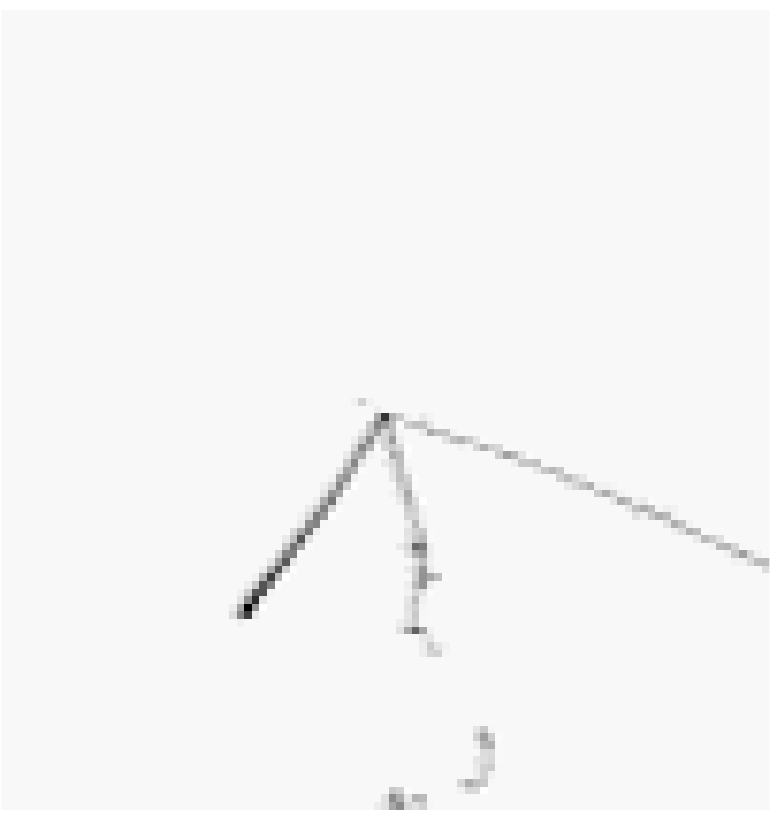}
\end{subfigure}
\begin{subfigure}[b]{0.298\textwidth}
	\caption{Polar, Coll, sig.}  
	\includegraphics[clip=true, width=\textwidth]{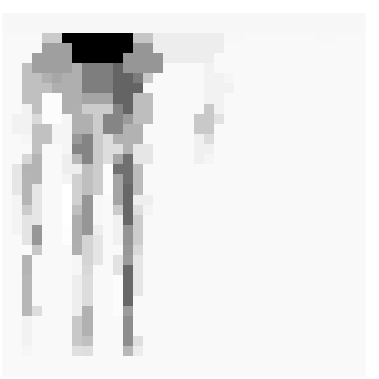}
\end{subfigure}
\begin{subfigure}[b]{0.39\textwidth}
	\caption{Distribution, Coll, sig.}  
	\includegraphics[clip=true, width=\textwidth]{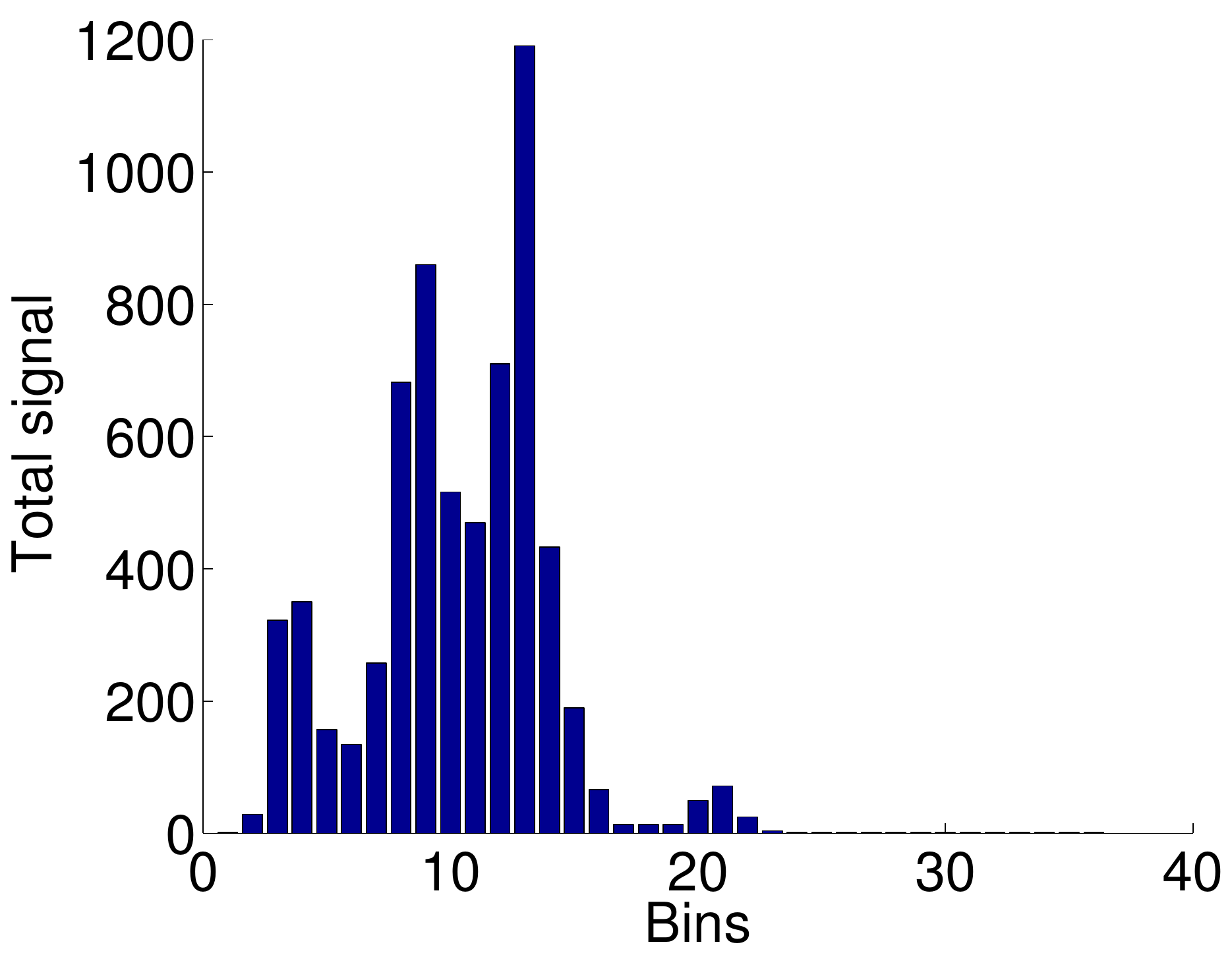}
\end{subfigure}
\caption{The example of negative (bkg.) and positive (sig.) event presented in different perspectives, namely as a raw image observed in the detector, the image in polar coordinates, and charge distribution with 36 bins and radius equal 10 pixels. The each event is presented in Induction2 (Ind2) and Collection (Coll) views. Each row describes one event in a selected view.} \label{distr}
\end{figure}

The events from different classes have charge amplitudes propagated in different ways starting from the PIV. Herein, the event's feature descriptor is proposed to describe this property. The event observed in the detector is described as two images, from Induction2 and Collection views. Each image is converted into the polar coordinate system with radius $R$ and number of bins $B$ spaced to each other with $360/B$ degrees and center in the PIV. The total charge in each bin is summed and creates a charge histogram for the considered view. Additionally, for each view the statistics variables are computed, which describe minimum, maximum, standard deviation, mean and total sum of charge in the histogram. The histogram values and statistics variables from both views form a feature vector. This results in $2(B+5)$ features describing each event. In the Fig. \ref{distr} are presented images of example events from positive and negative class, with corresponding images after conversion into polar coordinates and charge  histograms. It can be observed that the histograms from the event with negative class (Fig.\ref{distr} c,f) have one peak. Whereas, histograms for positive class (Fig.\ref{distr} i,l) have more than one peak. This is the main difference between positive and negative classes. What is more, the tracks in images from negative class events have broader peaks in the histograms, contrary to tracks from positive class events which appear as lines in the image and narrow peaks in the histograms. The classifier algorithm is learned to distinguish these properties coded in the feature vector.

\subsection{Classification framework}

In the proposed approach, the created feature descriptor of the event is an input for the classifier, which response is a probability whether image represents interaction of electron neutrino. The Random Forest \cite{breiman} algorithm was used as classifier. To asses the classifier's performance the Receiver Operating Characteristic (ROC) \cite{elements} curve was used, where 
\begin{equation}
\textrm{True Positive Rate} = \frac{TP}{TP+FN},  
\end{equation}
\begin{equation}
\textrm{False Positive Rate} = \frac{FP}{FP+TN}.
\end{equation}
The TP stands for true positives - correctly classified positive samples, TN are true negative - properly classified negative samples, FP are false positives - negative samples incorrectly classified, and FN are false negatives, which are positive samples improperly classified as negatives. Additionally, the Area Under ROC Curve (AUC) and accuracy (the number of all correctly classified samples) was used. They were computed on 5-fold cross-validation (CV) repeated 10 times for  stability of the obtained results.

\section{Results}

\begin{figure}[H]
\captionsetup[subfigure]{aboveskip=-11pt,belowskip=-13pt}
\centering
\includegraphics[clip=true, width=\textwidth]{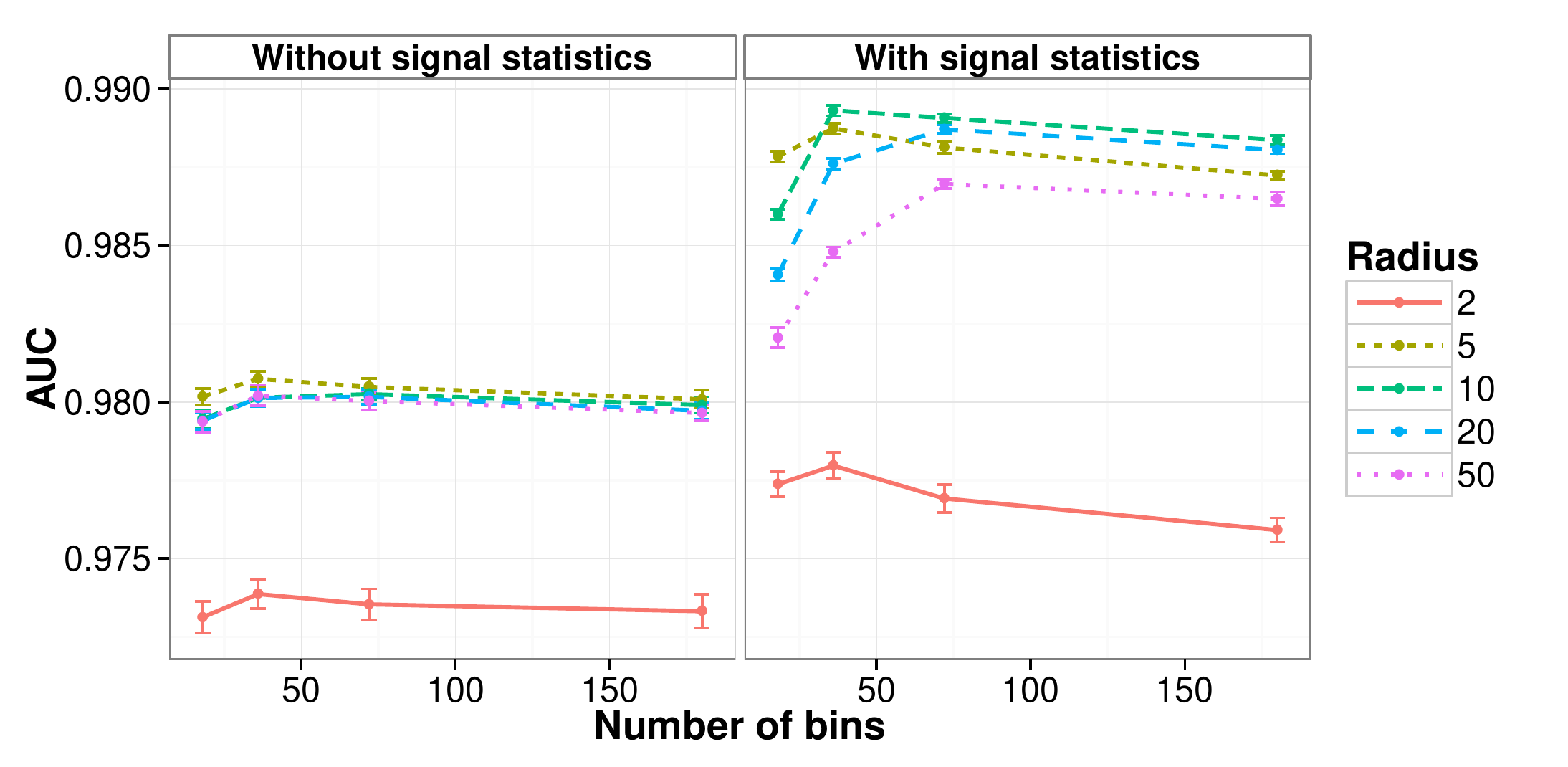}
\caption{The performance of the method computed on 5-fold CV repeated 10 times for different settings used for event's feature descriptor construction. There were used different number of bins, various length of the radius and presence of additional variables with signal statistics.} \label{diff_setting}
\end{figure}

The feature vector depends on two parameters: the number of bins and the length of the radius. In order to construct the most discriminative feature vector, the various parameters combinations were checked. There were considered a number of bins: $\{18, 36, 72, 180\}$, radius length: $\{2,5,10,20,50\}$ pixels and presence of signal statistics. To asses the discriminative power of feature vector, performance of the Random Forest (RF) classifier with 1000 trees was measured. The results are presented in the Fig.\ref{diff_setting}. There can be observed that performance of the classifier is higher when statistics variables are included in the feature vector. The best performance was obtained for the feature vector with 36 bins and radius length 10 pixels and signal statistics included, the AUC is 0.9893$\pm$0.0001 and accuracy is equal 0.9535$\pm$0.0007. From this point, in further experiments the feature vector is constructed with 36 bins, radius length equal 10 pixels and included signal statistics variables.


\begin{figure}
\vspace*{-7mm}
\centering
\includegraphics[clip=true, width=0.8\textwidth]{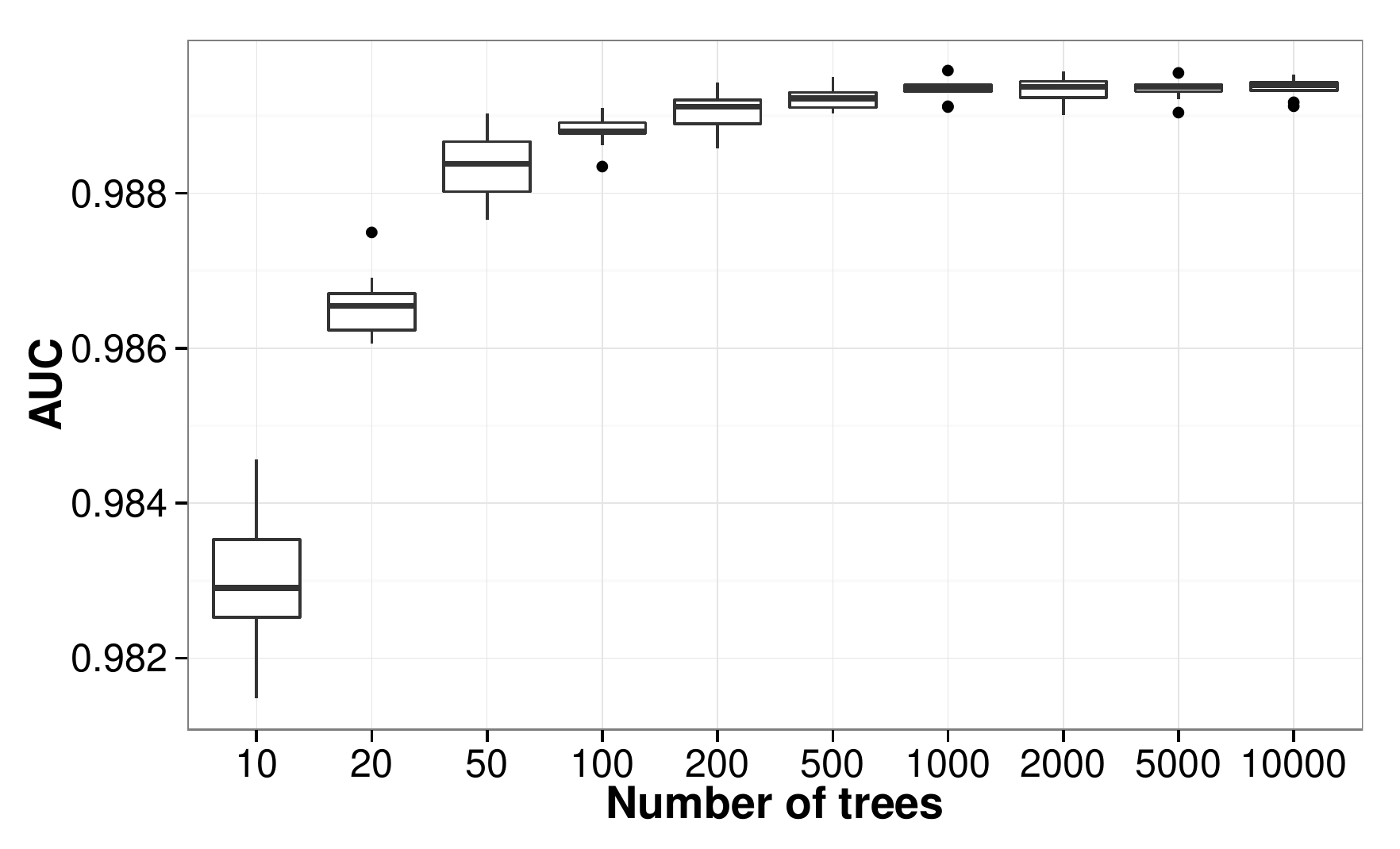}
\caption{The performance of the method computed on 5 fold CV repeated 10 times for different tree numbers for 36 bins and radius length equal 10 pixels and signal statistics included in feature vector.} \label{diff_trees}
\end{figure}

The performance of the proposed method in the dependency to the number of trees used in the RF classifier is presented in the Fig.\ref{diff_trees}. The AUC of the method increases with increasing tree number in the RF. However, the performance for 1000 or more trees in the forest is almost stable. In further analysis the 1000 trees in the RF were used. 

\begin{figure}[H]
\vspace*{-2mm}
\captionsetup[subfigure]{aboveskip=-11pt,belowskip=-13pt}
\centering
\begin{subfigure}[b]{0.49\textwidth}
	\caption{Dependency to noise in PIV} \label{noise_roc_1000} 
	\includegraphics[clip=true, width=\textwidth]{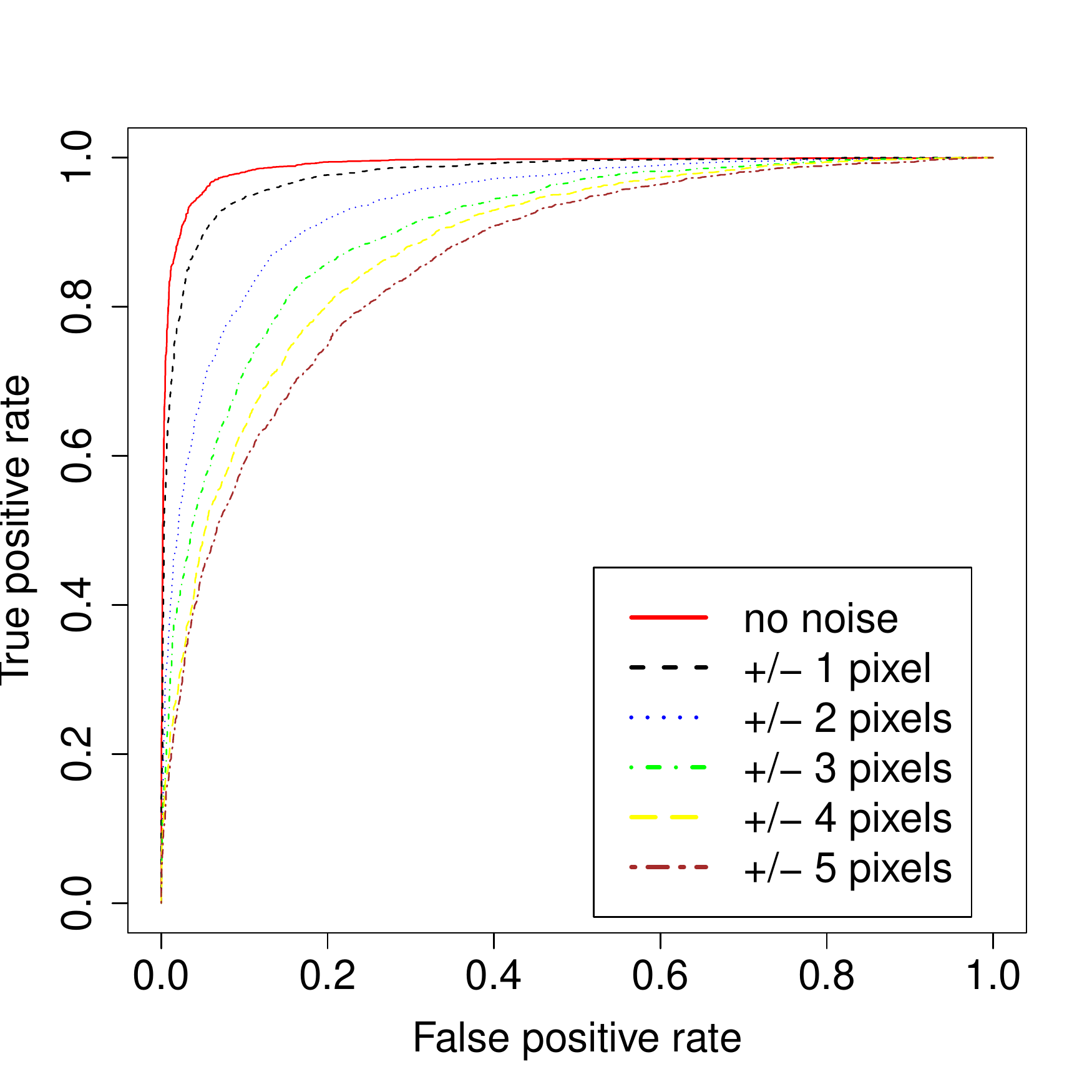}\end{subfigure}
\begin{subfigure}[b]{0.49\textwidth}
	\caption{Performance in energy ranges}  \label{energy_roc_1000} 
	\includegraphics[clip=true, width=\textwidth]{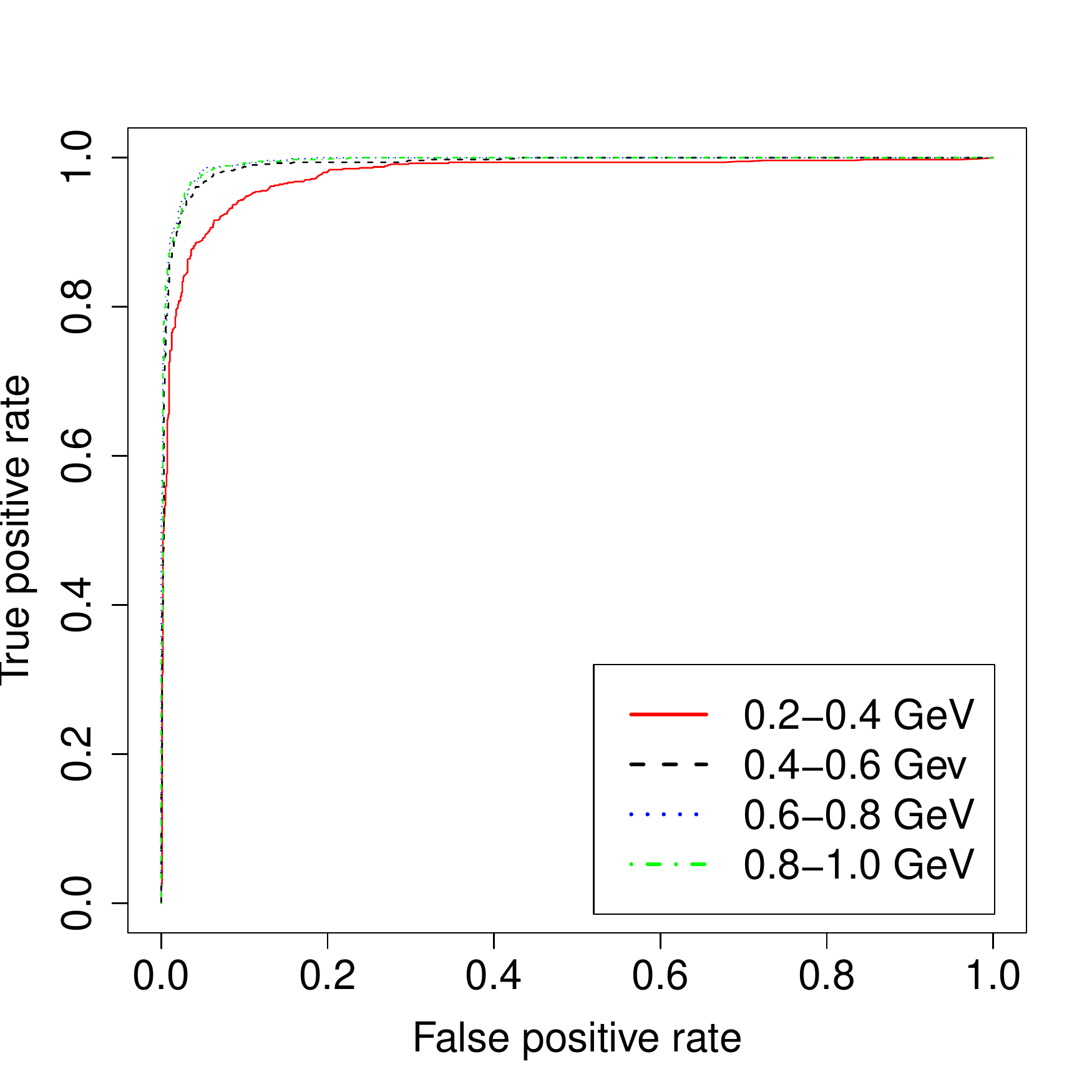}
\end{subfigure}
\caption{The performance of the proposed method for different (a) noise levels in PIV's location and (b) different energy of observed event.} \label{rocs}
\end{figure}

The proposed method assumes that PIV position is known. It should be designated before classification by another algorithm, for instance with algorithm presented in \cite{points_ben} and additional logic rules about PIV position. Therefore, the performance of the method for different noise levels in PIV's position is examined. The noise levels were generated by drawing a random number of pixels and adding to the true PIV's location. The ROC curves for different noise levels in PIV are presented in the Fig.\ref{noise_roc_1000} and the AUC values for 5-fold CV repeated 10 times are in Table \ref{noise_perf}. The performance of the method decreases with more noise in PIV's location. However, the method's AUC is 0.9360 for $\pm$ 2 pixels noise in PIV's location, which is considered as expected upper bound noise level of the algorithm which designates the PIV's location.

The image content depends on the observed event's energy. Therefore, the method performance were tested for different event's energy ranges (0.2-0.4, 0.4-0.6, 0.6-0.8, 0.8-1.0) GeV. The ROC curves are presented in the Fig.\ref{energy_roc_1000} and AUC values for 5-fold CV with 10 times repetition are in Table \ref{energy_perf}. It can be observed that for very low energies (0.2-0.4) GeV the performance of the method decreases slightly. However, for events with energy greater than 0.4 GeV the accuracy of the method is almost stable.
 
\begin{table}[ht]
\centering
\begin{tabular}{rr}
  \hline
 Noise level & AUC  \\ \hline
 zero noise & \ \ 0.9893 $\pm$ 0.0001\\
   1 pixel \  & 0.9771 $\pm$ 0.0004\\
   2 pixels & 0.9360 $\pm$ 0.0006\\
   3 pixels & 0.9031 $\pm$ 0.0009\\
   4 pixels & 0.8800 $\pm$ 0.0009\\
   5 pixels & 0.8601 $\pm$ 0.0014\\ \hline
\end{tabular}
\caption{The performance of the proposed method for different noise levels in PIV's location computed with 5-fold CV repeated 10 times.} \label{noise_perf}
\end{table} 

\begin{table}[ht]
\vspace*{-8mm}
\centering
\begin{tabular}{rr}
  \hline
 Energy range & AUC \\ \hline
 0.2 - 0.4 & \ \ 0.9770 $\pm$ 0.0005\\
 0.4 - 0.6 & 0.9906 $\pm$ 0.0002\\
 0.6 - 0.8 & 0.9944 $\pm$ 0.0003\\
 0.8 - 1.0 & 0.9934 $\pm$ 0.0002\\ \hline
\end{tabular}
\caption{The performance of the proposed method for different energy of observed event computed with 5-fold CV repeated 10 times.} \label{energy_perf}
\end{table}



\section{Conclusions}

The fundamental requirement in the case of surface neutrino detectors is an ability for rejection of cosmogenic background events. Herein, the novel method for classification of neutrino with electron flavour based on raw images from LAr-TPC detector is presented. The method constructs the feature vector for an observed event. It describes the distribution of the charge starting from primary interest vertex, which position is assumed to be known. The classifier makes a decision whether the detected event is an electron neutrino based on a feature descriptor. The best combination of parameters used for a feature vector construction was selected. The method has AUC 0.9893 and accuracy 0.9535. The experiments for different noise levels in PIV locations show that even with large noise (2 pixels, which corresponds to 2 wires on x-axis and 10 samples on y-axis randomly added in any direction) the AUC of the method is 0.9360. The performance of the method is slightly lower for low energy events ($<$ 0.4 GeV) and for events with energy higher than 0.4 GeV is almost constant with AUC greater than 0.99. The future work will focus on combining the proposed method with algorithm for PIV position estimation.

\section{Acknowledgements}
PP and KZ acknowledge the support of the National Science Center (Harmonia 2012/04/M/ST2/00775). Authors are grateful to the ICARUS Collaboration and Polish Neutrino Group for useful suggestions and constructive discussions during a preliminary part of this work.

\end{document}